%% file: neurips_2023.tex
\documentclass{article}

\usepackage[nonatbib,preprint]{neurips_2023}

\usepackage[utf8]{inputenc} 
\usepackage[T1]{fontenc}    
\usepackage{hyperref}       
\usepackage{url}            
\usepackage{booktabs}       
\usepackage{amsfonts}       
\usepackage{nicefrac}       
\usepackage{microtype}      
\usepackage{xcolor}         
\usepackage{multirow}

\usepackage{amsmath}
\usepackage{amssymb}
\usepackage{mathtools}
\usepackage{amsthm}

\usepackage{amsfonts}       
\usepackage{nicefrac}       
\usepackage{microtype}      
\usepackage{comment}
\usepackage{enumitem}

\usepackage{colortbl}
\usepackage{graphicx,color}
\usepackage{clrscode}
\usepackage{subfigure}
\usepackage{adjustbox}
\usepackage{multirow}
\usepackage{longtable}
\usepackage{clrscode}
\usepackage{array}
\usepackage{wrapfig}
\usepackage{enumitem}

\newcounter{noteZZctr} 

\usepackage{array}
\usepackage{arydshln}
\title{Beyond Text: A Deep Dive into Large Language Models' Ability on Understanding Graph Data}

\author{\parbox{0.9\linewidth}{
\centering{Yuntong Hu$^{*}$, {Zheng Zhang\thanks{These authors contributed equally to this work.}} ~, Liang Zhao 
} \\
{\rm Department of Computer Science\\
   Emory University \\
   Atlanta, GA 30322, USA }\\
 \texttt{\{yuntong.hu,zheng.zhang,liang.zhao\}@emory.edu} \\
}
}

\begin{document}

\maketitle

\begin{abstract}
  Large language models (LLMs) have achieved impressive performance on many natural language processing tasks. However, their capabilities on graph-structured data remain relatively unexplored. In this paper, we conduct a series of experiments benchmarking leading LLMs on diverse graph prediction tasks spanning node, edge, and graph levels. We aim to assess whether LLMs can effectively process graph data and leverage topological structures to enhance performance, compared to specialized graph neural networks. Through varied prompt formatting and task/dataset selection, we analyze how well LLMs can interpret and utilize graph structures. By comparing LLMs' performance with specialized graph models, we offer insights into the strengths and limitations of employing LLMs for graph analytics. Our findings provide insights into LLMs' capabilities and suggest avenues for further exploration in applying them to graph analytics.
\end{abstract}

\input{introduction}
\input{related_works}
\input{exp}
\input{conclusion}

\bibliographystyle{plain}
\bibliography{custom}

\end{document}

%% file: introduction.tex
\section{Introduction}

In recent years, there have been unprecedented advancements in large language models (LLMs)~\cite{qiu2020pre, ling2023domain} such as Transformers~\cite{Vaswani2017AttentionIA}, BERT~\cite{devlin-etal-2019-bert}, GPT~\cite{Brown2020LanguageMA}, and their variants. LLMs can be treated as foundation models that can be readily applied to diverse downstream tasks with little adaptation~\cite{Brown2020LanguageMA,kojima2022large,ling2023domain}. These models have achieved state-of-the-art results on many natural language processing tasks including text classification, machine translation, sentiment analysis, and text summarization~\cite{zhao2023survey}.
Significantly, advancements in architectures and training methodologies have given rise to emergent capabilities, setting state-of-the-art models like GPT-3.5~\cite{Brown2020LanguageMA}, GPT-4~\cite{OpenAI2023GPT4TR}, Claude-2~\cite{bai2022training}, BARD~\cite{bard2022}, LlaMA~{\cite{Touvron2023LLaMAOA}}, and LlaMA-2~{\cite{Touvron2023Llama2O}} apart from their predecessors. For instance, in-context learning~\cite{min2021metaicl} and zero-shot capabilities~\cite{kojima2022large, wei2021finetuned} enable these models to generalize across tasks for which they were not explicitly trained. This is confirmed by their excellent performance in complex activities such as mathematical reasoning and Question Answering (QA) systems. 

However, most of the tasks that Large Language Models (LLMs) surpassed previous benchmarks are Natural Language Processing (NLP) tasks involving sequential data. Graph-structured data presents additional complexity beyond sequences as it contains rich topological connections between entities that must be modeled along with node, edge, and graph attributes. Graph-structured data is ubiquitous across many domains, including social networks~\cite{newman2002random, ling2023deep}, knowledge graphs~\cite{paulheim2017knowledge}, molecular structures~\cite{wu2018moleculenet,zhang2021representation, wang2022multi}, and transportation networks~\cite{banavar1999size, zhang2022unsupervised}. While LLMs have shown powerful reasoning and generalization capabilities in sequential data, it remains unclear if they can handle structural information beyond context when applied to graph-structured data. This raises a compelling research question: Can the strengths of LLMs be extended to graph-structured data, enabling them to exhibit significant predictive ability? Further, can they compete with state-of-the-art models specialized for graph data, such as Graph Neural Networks (GNNs)?

To comprehensively study the capabilities of LLMs on graph-structured data, we conduct a series of empirical experiments with leading LLMs on diverse graph-based tasks that span node-, edge-, and graph-level predictions. By comparing their performance to specialized graph models like GNNs, we aim to assess the potential strengths and limitations of LLMs in this domain. Critically, by altering the input prompt formats, we aim to evaluate how effectively LLMs can extract and leverage the underlying structural information from the graph to enhance their performance in subsequent tasks. Additionally, we explore the importance of the structural data across different task dimensions spanning node, edge, and graph levels as well as diverse dataset domains such as citation networks, social networks, and chemical networks.

Broadly, this paper focuses on studying the central question of investigating the capabilities of LLMs on graph-structured data from three perspectives:
\begin{itemize}[leftmargin=*]
    \item \textbf{Can LLMs effectively process graph analytics tasks even without explicit graph structure?} Given that LLMs have already shown the capability to leverage contextual information for human-like reasoning in many NLP tasks, it becomes intriguing to assess whether they can attain substantial predictive performance on graph data tasks, even in the absence of structural information.
    \item \textbf{How well can LLMs interpret graph structures to enhance downstream task performance?} It is essential to investigate to what extent LLMs can perceive and interpret important graph structures. Furthermore, it is imperative to understand whether such recognition can influence and enhance performance in subsequent tasks.
    \item \textbf{How do task dimensions and dataset domains affect LLMs' ability to handle structured data?} LLMs' ability in identifying pivotal structural information for predictions can be influenced by specific tasks and data domains. For example, node-level tasks may heavily rely on entity attribute interpretation, while graph-level tasks may demand comprehensive understanding of intricate inter-node interations. Also, the distinct topologies properties to various dataset domains, whether derived from intricate social networks or sophisticated molecular structures, further influence the proficiency with which LLMs decipher and manage structured data.
\end{itemize}

The subsequent sections of this paper are structured as follows: We initiate with an extensive literature review, highlighting the recent advancements of LLMs within graph domains. Subsequent to this, we present our comprehensive findings on benchmarking LLMs on graph data, aiming to address the aforementioned research questions. This is accompanied by a detailed discussion, delving into the depth of our discoveries across varied experimental setups. We conclude by summarizing the key points and proposing ideas for future explorations.

%% file: related_works.tex
\section{Related Works}
\paragraph{Large language models for graph-structured data.} 
In recent literature, a few preliminary studies~\cite{ye2023natural,chen2023exploring,wei2023unleashing, guo2023gpt4graph} have made attempts to uncover the potential of LLMs in handling graph-structured data. Unfortunately, a comprehensive examination of LLMs' capacity to extract and harness crucial topological structures across diverse prompt settings, task levels, and datasets remains underexplored. Both Chen et al.\cite{chen2023exploring} and Guo et al.\cite{guo2023gpt4graph} proposed to apply LLMs directly on graph data. Their research primarily focus on the node classification task, constrained to a selected few datasets within the citation network domain, and thereby fails to offer a thorough exploration of LLMs' ability over diverse task levels and datasets. In addition, Ye et al.\cite{ye2023natural} fine-tuned LLMs on a designated dataset to outperform GNN, underscoring a distinct research objective compared to our study which emphasizes the intrinsic proficiency of LLMs in understanding and exploiting graph structures. Meanwhile, Wei et al.\cite{wei2023unleashing} treated LLMs as autonomous agents within graph data, which is less relevant to the core focus of our paper. 

\paragraph{Graph neural networks.}  In recent years, graph neural networks (GNNs)~\cite{kipf2016semi, defferrard2016convolutional, niepert2016learning,gilmer2017neural, hamilton2017inductive, xu2018powerful, morris2019weisfeiler,zhang2021representation,ling2023deep, chai2022distributed,wang2022deep, bai2023staleness} have emerged as a powerful deep learning approach for graph analysis and learning. GNNs operate by propagating information along edges of the graph and aggregating neighborhood representations for each node. The expressive power of GNNs to learn from graph structure makes them well-suited for analyzing complex relational data~\cite{xu2018powerful, zhu2020beyond,liu2020towards}. Unlike standard deep neural networks which operate on regular grids, GNNs can leverage the topological structure of graphs and have achieved state-of-the-art performance on tasks such as node classification~\cite{kipf2016semi}, link prediction~\cite{kumar2020link}, and graph classification~\cite{errica2019fair}.  

%% file: exp.tex
\section{Experiments}

\paragraph{Datasets.} We conducted the experiments on 5 commonly used graph benchmark datasets for node-level, edge-level and graph-level tasks: \textsc{Cora}~\cite{sen:aimag08}, \textsc{Pubmed}~\cite{sen:aimag08}, \textsc{OGBN-arxiv}~\cite{hu2020open}, \textsc{WordNet18}~\cite{miller1995wordnet} and \textsc{Reddit}~\cite{hamilton2017inductive}. Brief descriptions of the datasets are shown in Table \ref{data}.

\begin{table}[b] \centering
\begin{tabular}{@{}lrrrr@{}}\toprule
\textbf{Dataset} & \textbf{\#Node} & \textbf{\#Edge} & \textbf{\#Task} & \textbf{Metric}\\ \midrule
\textsc{Cora} & 2,708 & 5,278 & 7-class node classifi. \&Link Prediction & Accuracy\\ 
\textsc{Pubmed} & 19,717 & 44,324  & 3-class node classifi. \&Link Prediction  & Accuracy\\ 
\textsc{OGBN-arxiv} &  169,343 & 1,166,243 & 40-class node classifi.  & Accuracy \\ \midrule
\textbf{Dataset} & \textbf{\#Entity} & \textbf{\#Relation} & \textbf{\#Task} & \textbf{Metric}\\ \midrule
\textsc{WordNet18} &  40,943 & 18  &  18-class link classifi.  & Accuracy \\ \midrule
\textbf{Dataset} & \textbf{\#Node} & \textbf{\#Subgraph} & \textbf{\#Task} & \textbf{Metric}\\ \midrule
\textsc{Reddit} & 3,848,330  & 29,651  & 70-class subgraph classifi.  & Accuracy \\ 
\bottomrule
\end{tabular}
\caption{Statistics of the datasets. For \textsc{Reddit}, it actually contains 29,651 subreddits (classes). Here we only randomly sampled 70 communities for graph classification task in each run.}
\label{data}
\end{table}

We selected these five datasets for our preliminary experiments due to their rich contextual information present in the attributes of nodes, edges, and graphs. Specifically, \textsc{Cora}, \textsc{Pubmed} and \textsc{OGBN-arxiv} are citation network, where each node represents a research paper while an edge between two nodes indicates that there is a citation relationship between them. Edge in \textsc{WordNet18} links two synsets that are regarded as nodes. \textsc{Reddit} came from Reddit posts, in which each node represents a post and two nodes are connected if the same user comments on two posts. The specifics regarding their textual features are as follows:
\begin{itemize}
    \item \textsc{Cora}: Each node represents a paper in the domain of Artificial Intelligence, containing the information about its title and abstract. Each paper belongs to one of the following 7 categories: [`Case\_Based', `Theory', `Genetic\_Algorithms', `Probabilistic\_Methods', `Neural\_Networks', `Rule\_Learning', `Reinforcement\_Learning']. An edge from one node to another indicates the first paper cited the second one.
    \item \textsc{Pubmed}: Each node represents a scientific publication from PubMed database pertaining to diabetes. The node textual information contains keywords from its abstract and text body. Each paper belongs to one of the following 3 categories: [`Diabetes Mellitus, Experimental', `Diabetes Mellitus Type 1', `Diabetes Mellitus Type 2']. An edge from one node to another indicates the first paper cited the second one.
    \item \textsc{OGBN-arxiv}: Each node represents a research paper, containing the information about its title and abstract. Each paper belongs to one of 40 categories on arxiv.cs such as `AI' (Artificial Intelligence). An edge leading from one node to another signifies that the first paper cites the second one.
    \item \textsc{WordNet18}: Each node represents a synset, containing a description. An edge between two nodes indicate their relation such as `furniture', `includes', or `bed'. Each edge belongs to one of 18 relationships.
    \item \textsc{Reddit}: Each node corresponds to a post made by a user, which contains descriptions or discussions about a particular topic. Each graph symbolizes a subreddit (or community), with affiliations to one out of 29,651 distinct communities, for instance, `math'.
\end{itemize}
\paragraph{Choices of LLMs.} We opted to utilize OpenAI's state-of-the-art models, GPT-3.5 (GPT) and GPT-4, via their API system, based on a balance between performance and cost considerations. We adopted GPT with the latest versions (\textit{gpt-3.5-turbo-16k} and \textit{gpt-4}) in experiments.

\paragraph{Implementation Details.} For node classification task, we follow the same train-test split of \textsc{Cora}, \textsc{Pubmed} and \textsc{OGBN-arxiv} as established in semi-supervised GNN methods~\cite{kipf2016semi,xu2018powerful}. For link prediction on \textsc{Cora}, \textsc{Pubmed} and  \textsc{WordNet18}, a random 15\% of the links from the graph and the same number of negative-edge node pairs are packed into the test sets. For graph classification, in each run, we randomly selected 70 communities. Experiments conducted on \textsc{WordNet18} and the retrieval test for \textsc{Cora} employed few-shot prompts. Conversely, all other experiments leveraged zero-shot prompts. We executed each experiment thrice, subsequently averaging the results."

\paragraph{Comparison GNN Methods.} On node-level tasks, we choose the semi-supervised results from Graph Neural Network (GNN) \cite{gnn}, Graph Convolutional Network (GCN) \cite{gcn} and Graph Attention Network (GAT) \cite{gat} to compare with performance from LLMs. On edge-level tasks, we consider Graph Auto-Encoder (GAE) \cite{gae}, Graph InfoClust (GIC) \cite{gic}. It is worth noting this is not an absolutely fair comparison. Since LLMs operate under zero-shot or few-shot settings, where GNNs require a training set for parameter optimization. Additionally, potential data leakage during the LLMs' training process remains a concern. However, these studies aim to offer a foundational understanding of LLMs' proficiency in understanding graph data structures and forecasting downstream tasks.


\begin{minipage}{\textwidth}
\begin{minipage}[t]{0.48\textwidth}
\makeatletter\def\@captype{table}
\centering
\begin{tabular}{lccc} \toprule
\textbf{\multirow{2}*{Model}} & \multicolumn{3}{c}{Node-level} \\
\cline{2-4} & \textsc{Cora} & \textsc{Pubmed} &  \textsc{Arxiv} \\ \midrule
\textbf{GCN-64$^{*}$} & 0.814 & 0.790 & 0.731 \\ \hdashline
\textbf{GAT} & \textbf{0.832} & 0.790 & \textbf{0.742} \\ \hdashline
\textbf{GNN} & 0.829 & 0.738 & 0.721 \\ \hline
\textbf{GPT-3.5} & 0.627 & 0.673 & 0.516 \\ \hdashline
\textbf{GPT-4} & 0.432 & 0.821 & 0.642 \\\hline
\textbf{GPT-3.5$^*$} & 0.647 & 0.712 & 0.509 \\ \hdashline
\textbf{GPT-4$^*$} & 0.531 & \textbf{0.833} & 0.673 \\\hline
\textbf{GPT-3.5$^\oplus$} & 0.656 & 0.704 & 0.445 \\ \hdashline
\textbf{GPT-4$^\oplus$} & - & 0.814 & - \\\hline
\textbf{GPT-3.5$^\bullet$} & 0.054 & - & - \\ \hdashline
\textbf{GPT-4$^\bullet$} & 0.047 & - & - \\
\bottomrule
\end{tabular}
\caption{Average accuracy on node classification tasks. [\textbf{No structure information}] \textbf{GPT-3.5} and \textbf{GPT-3.5$^*$} mean zero-shot and few-shot prompt strategy. [\textbf{With structure information}] \textbf{GPT-3.5$^\oplus$} and \textbf{GPT-3.5$^\bullet$} mean prompts contain neighbors' information and retrieval requires, respectively.}
\label{node}
\end{minipage}\hspace{3mm}
\begin{minipage}[t]{0.48\textwidth}
\makeatletter\def\@captype{table}
\centering
\begin{tabular}{lccc} \toprule
\textbf{\multirow{2}*{Model}} & \multicolumn{3}{c}{Edge-level} \\
\cline{2-4} &  \textsc{Cora} & \textsc{Pubmed} & \textsc{WordNet} \\ \midrule
\textbf{GAE} & 0.793 & \textbf{0.923} & - \\ \hdashline
\textbf{GIC} & 0.812 & 0.775 & - \\ \hdashline
\textbf{GNN} & 0.739 & 0.528 & - \\ \hline
\textbf{GPT-3.5} & 0.512 & 0.116& 0.134\\ \hdashline
\textbf{GPT-4}& 0.578 & 0.132 & 0.169\\ \hline
\textbf{GPT-3.5$^\circ$} & 0.646 & 0.114 & -\\ \hdashline
\textbf{GPT-4$^\circ$}& \textbf{0.967} & 0.143 & -\\
\bottomrule
\end{tabular}
\caption{Average accuracy on link prediction tasks. \textbf{GPT-3.5$^\circ$} means adding structural information into prompt like Table \ref{prompt}. There are only triples for entries in \textsc{WordNet18}, which makes there is no connected graph structure for it.}
\label{edge}
\end{minipage}
\end{minipage}

\begin{table}[h] \centering
\begin{tabular}{@{}lccccccccccc@{}}\toprule
\textbf{\multirow{2}*{Model}} & \multicolumn{10}{c}{Number of labels} \\
\cline{2-11} & 1 & 5 &  10 & 15 & 20 & 30 & 40 & 50 & 60 & 70\\ \midrule
\textbf{GPT-3.5} & 0.773 & 0.662 & 0.618 & 0.594 & 0.628 & 0.604 & 0.638 & 0.536 & 0.618 & 0.507 \\ \hdashline
\textbf{GPT-4} & 0.957 & 0.886 &0.895 & 0.843 & 0.795 & - & - & - & - & - \\
\bottomrule
\end{tabular}
\caption{Average performance of GPT-3.5 and GPT-4 on \textsc{Reddit} when possible labels increase from 1 to 70. Results on GPT-4 with more labels are not available due to input limit of prompt length.}
\label{graph}
\end{table}


\subsection{Node-level task}
Driven by the goal of investigating LLMs' capabilities in discerning patterns within textual graphs and leveraging this for downstream tasks, we crafted three distinct prompts for our node-level prediction task experiments: (1) absence of graph topology descriptions; (2) straightforward presentation of all neighborhood data to the LLM; and (3) a retrieval-based prompt guiding the LLM to extract task-centric structural details. Examples of these prompts can be found in Table \ref{prompt}.

\textbf{LLMs' zero-shot or few-shot ability on node classificaiton tasks is still usually weaker than the semi-supervised GNN performance}. This may arguably suggests that general LLMs still can not surpass the specialized graph models on node classification task. As indicated in Table \ref{node}, GPT-4 outperforms the GNN models in zero-shot and few-shot settings on \textsc{Pubmed}, but this superiority isn't observed on \textsc{Cora} and \textsc{OGBN-arxiv}. We hypothesize three potential reasons for this discrepancy: 1. Fewer categories; 2. Less semantic overlap between categories; 3. Questionable groundtruth categories. GPT's `wrong' predictions about citation networks are mostly reasonable. Particularly, papers in \textsc{OGBN-arxiv} with lable of Computation and Language always are classified into other categories like Artificial Intelligence and Machine Learning. These prediction error papers always mentioned many machine learning concepts in their abstracts. We also argued whether datasets are `out of fashion' compared with the information in the GPT's corpus. We prompted GPT to use pre-2000 categorization criteria on \textsc{Cora}, but this does not lead to improvements. Intriguingly, GPT-4 did not consistently surpass GPT-3.5 in terms of predictive accuracy, hinting that the extent of pre-trained knowledge could significantly influence predictions in zero-shot scenarios.

\textbf{Incorporating structural information can slightly enhance the performance of GPT on node-level tasks to a certain degree.}  As evidenced in Table~\ref{node}, the inclusion of neighborhood information enhances node classification performance in certain instances. However, this improvement lacks consistency across different LLM selections and datasets. Such observations could suggest that structural information might not be a pivotal factor in node-classification tasks. Additionally, we assessed the capability of GPT to retrieve information by incorporating retrieval requirements into the prompts for \textsc{Cora}. Regrettably, this led to both GPT-3.5 and GPT-4 struggling significantly, rendering them largely unable to provide accurate predictions.

\begin{table}[t] \centering
\begin{tabular}{@{}lll@{}}\toprule
Node-level Task & Structure? & Prompt to GPT\\ \midrule
\textbf{Zero-shot} & No &  The title of one paper is \textcolor{blue}{<Title>} and its abstract is \textcolor{blue}{<Abstract>}.\\
(\textsc{Cora} \& \textsc{Pubmed}& & Here are possible categories: \textcolor{blue}{<Categories>}. Which category\\
\& \textsc{OGBN-arxiv}) & &  does this paper belong to? You can only output one category.  \\ \hdashline
 & Yes &  The title of one paper is \textcolor{blue}{<Title>} and its abstract is \textcolor{blue}{<Abstract>}.\\
& & This paper cited following papers: \textcolor{blue}{<TitleList>} and abstracts of \\
&&these papers are \textcolor{blue}{<AbstractList>}. Here are possible categories: \\
&&\textcolor{blue}{<Categories>}. Which category does this paper belong to? You\\
& & can only output one category. \\ \hdashline
 & Yes & The title of one paper is \textcolor{blue}{<Title>} and its abstract is \textcolor{blue}{<Abstract>}. \\
& & This paper is cited by following papers: \textcolor{blue}{<TitleList1>}. Each of\\
&& these papers belongs to one categories in: \textcolor{blue}{<Categories>}. You \\
&&need to 1.Analyse the papers' topic based on the give title and\\
&& abstract; 2.Analyse the pattern of citation information based on\\
&& on their titles, and retrieve the citation information you think is \\
&& important to help you determine the category of the first given\\
&& paper. Now you need to combine the information from 1 and 2\\
&& to predict the category of the first given paper. You should only \\
& & output one category. \\ \hline
\textbf{Few-shot} & Yes & Here is a list of labeled papers: The title and abstract of the first\\
&& paper are \textcolor{blue}{<Title1>} and \textcolor{blue}{<Abstract1>} respectively, and this pape\\
&&r belongs to \textcolor{blue}{<Category1>}$\cdots$ Here is a new paper whose title is \\
& & \textcolor{blue}{<Title>} and its abstract is \textcolor{blue}{<Abstract>}. Here are possible catego- \\
&&ries: \textcolor{blue}{<Categories>}. Which category does this paper belong to? \\
& & You can only output one category. \\ 
\bottomrule
\end{tabular}
\caption{Examples of different prompts used in node classification experiments. For few-shot tasks, we randomly sampled two papers for each category due to the limit of input length. }
\label{prompt}
\vspace{-6mm}
\end{table}

\subsection{Edge-level task}
\textbf{Contrary to node-level tasks, the structural information of a graph seems to be more crucial for edge-level tasks}. When only node data, excluding neighborhood information, is presented to GPT, the link prediction accuracy on \textsc{Cora} stands at 51.2\% for GPT-3.5 and 57.8\% for GPT-4. Remarkably, these figures signicantly increased to 64.3\% and 96.7\% respectively when we incorporate the nodes' neighbors information. Notably, the zero-shot result of GPT-4 even surpass the performances of all trained GNN models. It is worth noting that some wrongly predicted links can be attributed to the absence of neighbor information for these nodes in the dataset. Table \ref{prompt} illustrates the difference between prompts used on link prediction tasks. It is also interesting that when we further increase the information of neighboring nodes (e.g., the abstract of an article), the prediction accuracy becomes worse than only with information of titles. For the link prediction task on \textsc{WordNet18}, we randomly selected 5 entries for each relationship and presented this information to GPT. Unfortunately, both GPT-3.5 and GPT-4 struggled to achieve a high predictive accuracy based on the provided data. A plausible explanation for this could be GPT's heavy reliance on its pre-trained knowledge, especially when not fine-tuned for specific tasks.

\begin{table}[t] \centering
\begin{tabular}{@{}lll@{}}\toprule
Edge-level Task & Structure? & Prompt to GPT\\ \midrule
\textbf{Zero-shot} & No &  There are two papers. Title of the first paper is \textcolor{blue}{<Title1>} and its \\
(\textsc{Cora} \& \textsc{Pubmed})& & abstract is \textcolor{blue}{<Abstract1>}. Title of the second paper is \textcolor{blue}{<Title2>}, \\
& &  and its abstract is \textcolor{blue}{<Abstract2>}. You need to predict whether the \\
& &  second paper or not. Answer 'YES' or 'NO'.\\ \hdashline
 & Yes &  There are two papers. Title of the first paper is \textcolor{blue}{<Title1>} and its \\
& & abstract is \textcolor{blue}{<Abstract1>}. Title of the second paper is \textcolor{blue}{<Title2>}, \\
& &  and its abstract is \textcolor{blue}{<Abstract2>}. The first paper cited following\\
&&  papers: \textcolor{blue}{<Titles>}. You need to predict whether the second paper \\
& & or not. Answer 'YES' or 'NO'.\\ \hline
\textbf{Few-shot} & No & We define two descriptions should have a relationship, such as  \\
(\textsc{WordNet18})&&furniture <includes> bed. There are some samples: \textcolor{blue}{<Entries>}.\\
& & Here are possible relations: \textcolor{blue}{<Relationships>}. The first entry is\\
&&\textcolor{blue}{<Entry1>} and the second entry is \textcolor{blue}{<Entry2>}. You must output \\
&&only one relationship between these two entries.\\
\bottomrule
\end{tabular}
\caption{Examples of different prompts used in link prediction experiments. Structural information plays an important role in link prediction task.}
\label{prompt}
\vspace{-5mm}
\end{table}

\vspace{-2mm}
\subsection{Graph-level task}\vspace{-1mm}
For graph-level tasks, we only conducted experiments on \textsc{Reddit} due to its text richness and semantic ambiguity. Only GPT-3.5 was tested since the information of one community is large even we have summarized the information from each user. We selected \textit{top-k} post summaries of the most replied users as representative information of a community. As shown in Table \ref{graph}, when GPT needs to make predictions from full 70 communities, the accuracy was 50.7\%. The accuracy decreased from 77.3\% to 50.7\% when possible communities in the \textcolor{blue}{<SubReddits>} list increased from 1 to 70.


\begin{table}[h] \centering
\begin{tabular}{@{}lll@{}}\toprule
Graph-level Task & Structure? & Prompt to GPT\\ \midrule
\textbf{Zero-shot} & Yes & There are texts from representative users of one Reddit community: \\
(\textsc{Reddit})&&\textcolor{blue}{<Posts>}. There are following communities: \textcolor{blue}{<SubReddits>}. Which\\
& & community does these texts belong to? You should only output one\\
&&community from given communities.\\
\bottomrule
\end{tabular}
\caption{Example prompt used in graph classification experiments. Structural information is given by a list of \textit{top-k} important nodes a graph.}
\label{graphprompt}
\vspace{-5mm}
\end{table}

%% file: conclusion.tex
\section{Conclusion and Future Works}
This research presented a systematic empirical evaluation of leading LLMs on diverse graph learning tasks spanning node, edge, and graph levels. Through careful variation of prompt design and dataset selection, we assessed the capabilities of models such as GPT-3.5 and GPT-4 in interpreting and leveraging graph structural information to enhance predictive performance. Our results demonstrate that while LLMs exhibit reasonable node classification capabilities even without explicit graph data, likely by relying on contextual clues, their zero-shot performance continues to lag behind state-of-the-art GNNs specialized for this domain. However, incorporating graph topology information can significantly boost performance on edge-level link prediction tasks, with GPT-4 even surpassing certain GNNs in select cases. On more complex graph classification tasks, limitations emerge in handling increased output complexity. In summary, this research provides valuable evidence that LLMs have promising capabilities on graph analytics, while also revealing clear areas for improvement compared to specialized graph models.

Our future work should explore more rigorous benchmarking LLMs on graph learning tasks with graph specialized models, novel prompt designs to focus on topological structures, evaluating on additional graph tasks, and even fine-tuning open-sourced LLMs on graphs. By exploring these avenues, the full potential of large language models for advancing graph representation learning and analytics can be more promising.